\documentclass[10pt,twocolumn,letterpaper]{article}

\usepackage{cvpr}              

\usepackage{graphicx}
\usepackage{amsmath}
\usepackage{amssymb}
\usepackage{booktabs}
\usepackage[ruled,vlined]{algorithm2e}
\usepackage{epsfig}
\usepackage{physics}
\usepackage[inline]{enumitem}
\usepackage{comment}
\usepackage{subcaption}
\usepackage{framed}
\usepackage{multirow}
\usepackage{xcolor,soul}
\usepackage{latexsym,url}
\usepackage{stackengine} 

\usepackage[pagebackref,breaklinks,colorlinks]{hyperref}

\usepackage[capitalize]{cleveref}
\crefname{section}{Sec.}{Secs.}
\Crefname{section}{Section}{Sections}
\Crefname{table}{Table}{Tables}
\crefname{table}{Tab.}{Tabs.}

\def\cvprPaperID{8360} 
\def\confName{CVPR}
\def\confYear{2022}

\definecolor{todocolor}{RGB}{200,120,120}
\sethlcolor{todocolor}

\def\etal{{et.~al}}
\def\sota{\textsc{sota}\xspace}
\def\cnn{\textsc{cnn}\xspace}
\def\usg{\textsc{usg}\xspace}
\def\gbc{\textsc{gbc}\xspace}
\def\gb{\textsc{gb}\xspace}
\def\mri{\textsc{mri}\xspace}
\def\ct{\textsc{ct}\xspace}
\def\roi{\textsc{roi}\xspace}
\def\rois{\textsc{roi}s\xspace}
\def\gbcnet{\textsc{gbcn}et\xspace}

\newcommand{\myfirstpara}[1]{\noindent \textbf{#1:}}
\newcommand{\mypara}[1]{\vspace{0.1em} \myfirstpara{#1}}

\newcommand\oast{\stackMath\mathbin{\stackinset{c}{0ex}{c}{0ex}{\ast}{\bigcirc}}}

\newcommand{\beginsupplement}{%
        \setcounter{table}{0}
        \renewcommand{\thetable}{S\arabic{table}}%
        \setcounter{figure}{0}
        \renewcommand{\thefigure}{S\arabic{figure}}%
     }

\begin{document}

\title{Surpassing the Human Accuracy: \\ Detecting Gallbladder Cancer from USG Images with Curriculum Learning}

\author{Soumen Basu\textsuperscript{1}, Mayank Gupta\textsuperscript{1}, Pratyaksha Rana\textsuperscript{2}, Pankaj Gupta\textsuperscript{2}, Chetan Arora\textsuperscript{1} \\
\textsuperscript{1} Indian Institute of Technology, Delhi, India \\ 
\textsuperscript{2} Postgraduate Institute of Medical Education and Research, Chandigarh, India\\
}

\maketitle

\begin{abstract}
We explore the potential of \cnn-based models for gallbladder cancer (\gbc) detection from ultrasound (\usg) images. \usg is the most common diagnostic modality for \gb diseases due to its low cost and accessibility. However, \usg images are challenging to analyze due to low image quality because of noise and varying viewpoints due to the handheld nature of the sensor. Our exhaustive study of state-of-the-art (\sota) image classification techniques for the problem reveals that they often fail to learn the salient \gb region due to the presence of shadows in the \usg images. \sota object detection techniques also achieve low accuracy because of spurious textures due to noise or adjacent organs. We propose GBCNet to tackle the challenges in our problem. GBCNet first extracts the regions of interest (\rois) by detecting the \gb (and not the cancer), and then uses a new multi-scale, second-order pooling architecture specializing in classifying \gbc. To effectively handle spurious textures, we propose a curriculum inspired by human visual acuity, which reduces the texture biases in GBCNet. Experimental results demonstrate that GBCNet significantly outperforms \sota \cnn models, as well as the expert radiologists. Our technical innovations are generic to other \usg image analysis tasks as well. Hence, as a validation, we also show the efficacy of GBCNet in detecting  breast cancer from \usg images. Project page with source code, trained models, and data is available at \href{https://gbc-iitd.github.io/gbcnet.html}{\texttt{https://gbc-iitd.github.io/gbcnet}}. 
\end{abstract}

\begin{figure}[t]
    \centering
    \includegraphics[width=\linewidth]{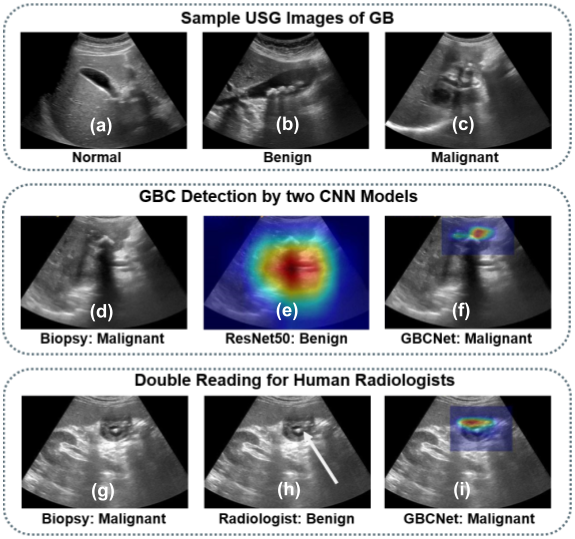}
    \caption{(a), (b), and (c) Normal, benign, and malignant \gb sample in \usg images, respectively. While normal or benign \gb have regular anatomy, clear boundary is absent in malignant \gb. (d) A malignant (biopsy-proven) \gb sample. (e) Shadows having visual traits of a \gb leads to localization error in ResNet50. (f) \gbcnet tackles shadow artifacts well. (g) Another sample of malignant \gb. (h) The radiologist incorrectly diagnosed the \gb as benign based on the stone and wall thickening. (i) \gbcnet helps the radiologist to identify the salient region with liver infiltration by the \gb, a critical feature of \gbc, and correct the prediction.}
    \label{fig:teaser}
\end{figure}

\section{Introduction}

According to GLOBOCAN 2018 \cite{bray2018global}, worldwide about 165,000 people die of \gbc annually. For most patients, \gbc is detected at an advanced stage, with a mean survival rate for patients with advanced \gbc of six months and a 5-year survival rate of 5\% \cite{randi2006gallbladder, gupta2021locally}. Detecting \gbc at an early stage could ameliorate the bleak survival rate. 

Lately, machine learning models based on convolutional neural network (\cnn) architectures have made transformational progress in radiology, and medical diagnosis for diseases such as breast cancer, lung cancer, pancreatic cancer, and melanoma \cite{ardila2019end, bejnordi2017diagnostic, chu2019application, codella2017deep, han2017breast}. However, their usage is conspicuously absent for the \gbc detection. Although there has been prior work involving segmentation and detection of the \gb abnormalities such as stones and polyps \cite{gbPolyp, gbPolyp2, gbAutomatic}, detection of \gbc is missing from the list. A search on Google Scholar with keywords ``artificial intelligence'' and ``gallbladder cancer'' returned 204 articles between 2015-2021 (November). In these, we did not find any published article on deep learning-based \gbc detection from \usg images.

 
Early diagnosis and resection are critical for improving the survival rate of \gbc. Due to the non-ionizing radiation, low cost, portability, and accessibility, \usg is a popular diagnostic imaging modality. Although identifying anomalies such as stones or \gb wall thickening at routine \usg is easy, accurate characterization is challenging \cite{gupta2020imaging, gb-rads-paper}. Often, \usg is the sole diagnostic imaging performed for patients with suspected \gb ailments. If malignancy is not suspected, no further testing is usually performed, and \gbc could silently advance. Therefore, it is imperative to develop and understand the characterization of \gb malignancy from \usg images.

There are significant challenges in using \cnn models for \usg image analysis. Unlike \mri or \ct, \usg images suffer from low imaging quality due to noise and other sensor artifacts. The views are also not aligned due to the handheld nature of the sensor. We observe that modern \cnn classifiers fail to localize the salient \gb region due to the presence of shadows which often have similar visual traits of a \gb in \usg images (\cref{fig:teaser}). Training object detectors for \gbc detection gets biased towards learning from spurious textures due to noise and adjacent organ tissues rather than the shape or boundary of \gb wall, which results in poor accuracy. Further, unlike normal and benign \gb regions, which have regular anatomy, malignant cases are much harder to detect due to the absence of a clear \gb boundary or shape and the presence of a mass.

\mypara{Contributions} 
The key contributions of this work are:
\vspace{-0.5em}
\begin{enumerate}[label=\textbf{(\arabic*)}]
\itemsep-0.5em
	\item We focus on circumventing the challenges for automated detection of \gbc from \usg images and propose a deep neural network, GBCNet, for detecting \gbc from \usg images. GBCNet extracts candidate regions of interest (\rois) from the \usg to mitigate the effects of shadows and then uses a new  multi-scale, second-order pooling-based (MS-SoP) classifier on the \rois to classify gallbladder malignancy.
	MS-SoP encodes rich feature representations for malignancy detection. 
	\item Even though GBCNet shows improvement in \gb malignancy detection over multiple \sota models, the spurious texture present in an \roi bias the classification unit towards generating false positives. To alleviate the issue, we propose a training curriculum inspired by human visual acuity \cite{kwon2016compensation, vogelsang2018VisualAcuity}. Visual acuity refers to the sharpness of visual stimuli. 
	The proposed curriculum mitigates texture bias and helps GBCNet focus on shape features important for accurate \gbc detection from \usg images.
	\item A lack of publicly available \usg image datasets related to \gb malignancy adds to the difficulty of utilizing \cnn models for detecting \gbc. We have collected, annotated, and curated a \usg image dataset of 1255 abdominal \usg images collected from 218 patients. We refer this dataset as the Gallbladder Cancer Ultrasound (GBCU) dataset.
\end{enumerate}

\begin{figure*}[t]
	\centering
	\includegraphics[width=\textwidth]{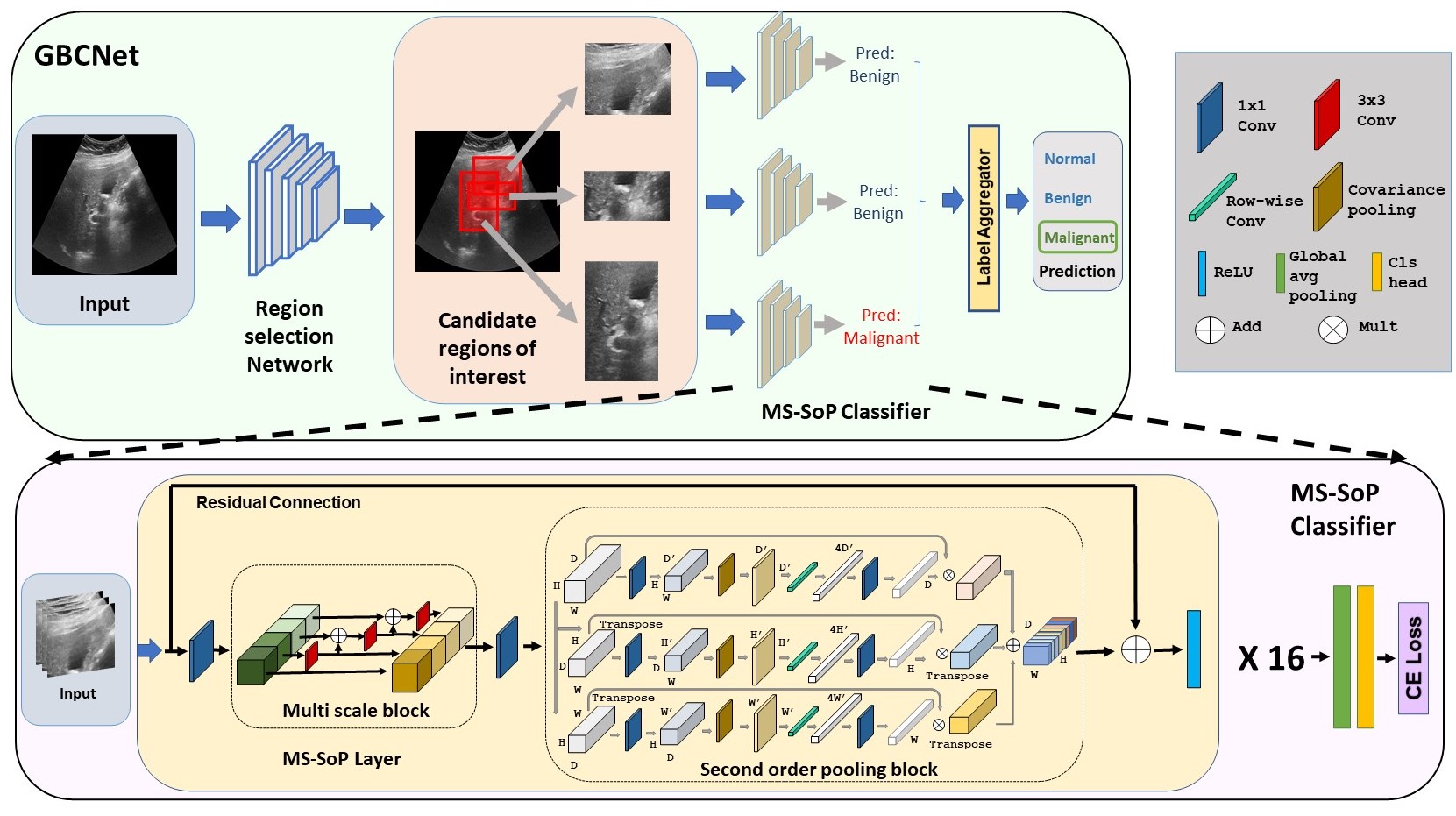}
	\caption{Overview of the \gbcnet architecture. The region selection network localizes the candidate regions of interest and the multi-scale, second-order pooling-based (MS-SoP) classifier at the next stage predicts malignancy for each region. The predictions for each region is aggregated to get the final prediction on the whole image.}
	\label{fig-arch-overview}
\end{figure*}

\section{Related Work}

\myfirstpara{Deep Learning for GB Abnormalities}
\usg imaging is an effective modality for diagnosing \gbc and related \gb afflictions \cite{yuan2018gbcManual}. 
Lien \etal \cite{gbAutomatic} use a parameter-adaptive pulse-coupled neural network for \gb stone segmentation in \usg images. Pang \etal \cite{gbYolo} identify \gb and gallstones using a YOLOv3 model from \ct images. 
Jeong \etal \cite{gbPolyp2} uses an InceptionV3 model to classify neoplastic polyps from cropped samples of \gb \usg images. \gbc is a serious problem affecting a significant number of people. Despite the presence of numerous studies on using deep learning on other \gb-related afflictions, there is an absence of any work that applies deep learning for \gbc detection on \usg images.

\mypara{Deep Learning for USG}
\cnn models have been widely applied in \usg imaging tasks, such as ovarian cancer detection \cite{zhang2019improved}, breast cancer region, mass and boundary detection \cite{bian2017boundary, cao2019BreastLesion, yap2018breast, ning2020multi, zhu2020second}, measuring head circumference in fetal \usg images \cite{sobhaninia2019FetalHC, budd2019FetalHC}. 


\mypara{Curriculum Learning}
Curriculum learning has been applied to different medical imaging tasks. While Jesson \etal \cite{jesson2017cased} used a patch-based curriculum for lung nodule detection, Tang \etal \cite{tang2018attention} used disease severity level to identify thoracic diseases from chest radiographs. Oksuz \etal \cite{oksuz2019automatic} proposed image corruption-based curriculum to detect motion artifacts in cardiac \mri. 

\mypara{Texture Bias in Neural Networks}
Presence of mass and a thickened \gb wall are prominent indicators of \gb abnormality. However, typical \cnn-based architectures are biased towards textures rather than shape \cite{geirhos2018Texture}. This may lead GBCNet to focus on soft tissue textures such as liver rather than noticing cues based on the shape and wall of the \gb. 
Multiple works have attempted to reduce texture bias and improve the spatial understanding of a model. Geirhos \etal \cite{geirhos2018Texture} suggest style transfer to replace the original texture of images while Brendel \etal \cite{brendel2019BagOfFeatures} propose a method similar to a Bag of Features model to force spatial learning. 

\mypara{Visual Acuity in Learning Models}
Vogelsang \etal \cite{vogelsang2018VisualAcuity} suggest that a period of low visual acuity (blurred vision) followed by high visual acuity induces better spatial processing and also increases the receptive field in human vision. 
Different from our visual acuity-inspired strategy of working with input space, Sinha \etal \cite{sinha2020curriculumBySmoothing} propose applying a Gaussian kernel on the output feature map of every layer of a network. The use of blurring before pooling seems to mitigate aliasing effects due to sub-sampling in the pooling layer rather than the use of visual acuity. Azad \etal \cite{azad2020textureDoG} have integrated a Difference of Gaussian (DoG) operation into their model. Similar to \cite{sinha2020curriculumBySmoothing}, they end up attenuating the high frequency in the feature maps corresponding to every layer rather than the input image, for which there is no obvious biological connection known. On the other hand, our proposed visual acuity-based curriculum works in the input space and has a solid neural basis \cite{vogelsang2018VisualAcuity}.

\section{Proposed Method}
\subsection{Model Architecture}
%
The artifacts in \usg images often result in multiple spurious areas in a \usg image with very similar visual traits as the \gb region, leading to a disappointing performance by the baseline classifiers. 
GBCNet selects candidate regions of interest (\rois) from the \usg to mitigate the influence of spurious artifacts like shadows and then uses a multi-scale, second-order pooling-based (MS-SoP) classifier on the \rois. \cref{fig-arch-overview} presents a conceptual diagram of the architecture. 
At test time, the detector may predict multiple \rois overlapping with the \gb. Occasionally, the detection network may fail to localize the \rois. In this case, we pass the entire image to the classifier. The proposed MS-SoP classifier exploits a range of spatial scales 
and second-order statistics to generate rich features from \usg images to learn the characteristics of malignancy. We run the classifier for each candidate region during inference and aggregate the predicted labels to compute the prediction for the entire image. If any of the \rois is classified as malignant, the image as the whole is classified as malignant. If all the regions are predicted to be normal, the image is classified as normal. In all other cases, the image is predicted to be benign. 

\mypara{Candidate \roi Selection}
We used deep object detection networks to localize the \gb region in a \usg image. The predicted bounding boxes serve as candidate regions of interest and mitigate the adverse effect of noise and artifacts from non-\gb regions during the classification. For training the \roi selection models, we use only two classes - background and the \gb region. In this stage, we only detect the \gb and do not classify them as malignant or non-malignant. In principle, it is possible to do both in a single stage, but we observed 
that using a separate classifier on the focused \rois leads to better accuracy. We note that our findings regarding the superiority of using classification on focused regions, instead of over the entire image, are consistent with other recent works \cite{sirinukunwattana2016locality, cao2019BreastLesion, lancet_pancreas, fan2020inf, eccv2020_devil_in_classification}. Prior studies demonstrate that modern object detection architectures such as YOLO \cite{yolov4} or Faster-RCNN \cite{fasterrcnn} can detect breast lesions in \usg images  \cite{cao2019BreastLesion}. On the other hand, recently proposed anchor-free approaches, such as Reppoints \cite{reppoints}, and CentripetalNet \cite{centripetalnet} can detect unconventionally sized objects such as \gb. Hence, we experimented with all the above approaches for \roi selection in our framework. 

\mypara{MS-SoP Classifier}
%
Modeling higher-order statistics has gained popularity in recent years due to its enhanced ability to capture complex features and non-linearity in deep neural networks \cite{zoumpourlis2017non, li2017second, gao2019global}. 
Ning \etal \cite{ning2020multi} recently used higher-order feature fusion for classifying breast lesions. They have used RGB image patches of three fixed scales at the input layer. We take the idea further and develop a novel multi-scale, second-order pooling (MS-SoP) layer to encode rich features suitable for malignant \gb detection. In contrast to \cite{ning2020multi}, we exploit feature maps of multiple scales in all the intermediate layers to learn a rich representation. The proposed MS-SoP layers can be conveniently plugged into any \cnn backbone.
The MS-SoP classifier contains $16$ MS-SoP layers as the backbone, followed by global average pooling and a fully connected classification head. 
We use the Categorical Cross-Entropy loss to train the classifier. 

\myfirstpara{Multi-Scale Block}
Abdominal organs can appear in significantly different sizes in a \usg image based on the insonation angle or the pressure on the transducer. Perceiving information across multiple scales is thus necessary for accurate \gbc detection. Recently, Gao \etal \cite{res2net} have replaced the standard convolution block in the bottleneck layer with group convolution to add a multi-scale capability to the ResNet architecture. Inspired by \cite{res2net}, we used a hierarchy of convolution kernels on slices of feature volumes in the intermediate layers to capture multi-scale information through a combination of different receptive fields. We split a feature map volume, $\mathcal{X} \!\in\!\mathbb{R}^{H\!\times\! W \!\times\! D}$ ($H, W ~\text{and}~D$ are the height, width, and the number of channels, respectively), depth-wise into 4 slices, $\mathcal{X}_1,\mathcal{X}_2,\mathcal{X}_3$, and $\mathcal{X}_4$, where $\mathcal{X}_i \!\in\! \mathbb{R}^{H\!\times\! W\!\times\! D/4}$. Each $\mathcal{X}_i$ will generate an output split $\mathcal{Y}_i$. The final output, $\mathcal{Y}$, is obtained by concatenating the splits. Suppose $\mathcal{C}_j$ are $3\!\times\!3$ convolution kernels and $\oast$ denotes the convolution. We get each $\mathcal{Y}_i$ as follows:
\linebreak
\begin{minipage}{.4\linewidth}
\vspace{-1em}
\begin{align}
    \mathcal{Y}_1 &= \mathcal{X}_1 \\
    \mathcal{Y}_2 &= \mathcal{C}_1 \oast \mathcal{X}_2
\end{align}
\end{minipage}
\begin{minipage}{.6\linewidth}
\vspace{-1em}
\begin{align}
    \mathcal{Y}_3 &= \mathcal{C}_2 \oast (\mathcal{X}_3 \!+\! \mathcal{Y}_2) \\
    \mathcal{Y}_4 &= \mathcal{C}_3 \oast (\mathcal{X}_4 \!+\! \mathcal{Y}_3)
\end{align}
\end{minipage}

\mypara{Second-order Pooling Block}
Traditional average or max-pooling use first-order statistics and thus cannot capture the higher-order statistical relation between features. Inspired by the recent success of higher-order statistics in breast lesion \usg \cite{ning2020multi, zhu2020second}, we employ the second-order pooling (SoP) mechanism to exploit the second-order statistical dependency between the multi-scale features.  

For computational efficiency, we reduce the number of channels of a feature volume, $\mathcal{X} \!\in\! \mathbb{R}^{H \!\times\! W \!\times\! D}$, to $D'~(D'\!<\!D)$, using $1\!\times\!1$ convolutions. $\mathcal{X}$ is then reshaped to a matrix $\vb{X} \!\in\! \mathbb{R}^{D'\!\times\! N}$, where $N\!=\!H\!\times\! W$. We compute the covariance of $\vb{X}$ as, $\vb{C}_{D'\!\times\!D'} \!=\! (\vb{X\!-\!\bar{X}})\vb{(X\!-\!\bar{X})^T}$, which is then reshaped to a tensor of size $1 \!\times\! D' \!\times\! D'$ and passed through a convolution layer with $4D'$ kernels of size $1 \!\times\! D'$ each. We resize the resulting $1\!\times\! 1 \!\times\! 4D'$ tensor to a $1\!\times\! 1 \!\times\! D$ tensor, $\vb{w_d}$, by $1\!\times\! 1$ convolutions. $\vb{w_d}$ represents the weight for each channel. These weights are then channel-wise multiplied with $\mathcal{X}$, to obtain the weighted feature map $\vb{Z_d}$. To repeat similar processes for the height and width, we transpose $\mathcal{X}$ from to a $D \!\times\! W \!\times\! H$ tensor, say $\mathcal{X}_h$ and to a $H \!\times\! D \!\times\! W$ tensor, say $\mathcal{X}_w$. In terms of index notation, $\mathcal{X}_h[k,j,i] \!=\! \mathcal{X}[i,j,k]$ and $\mathcal{X}_w[i,k,j] \!=\! \mathcal{X}[i,j,k]$, where $i\!=\!\{1,2,\ldots H\}, ~j\!=\!\{1,2,\ldots W\},$ and $k\!=\!\{1,2,\ldots D\}$. Similar to calculating $\vb{w_h}$ from $\mathcal{X}$, we find $\vb{w_h}\!\in\! \mathbb{R}^{1\!\times\!1\!\times\!H}$ from $\mathcal{X}_h$, and $\vb{w_w}\!\in\! \mathbb{R}^{1\!\times\!1\!\times\!W}$ from $\mathcal{X}_w$. We also calculate $\vb{Z_h}$ and $\vb{Z_w}$ by multiplying $\vb{w_h}$ and $\vb{w_w}$ channel-wise to $\mathcal{X}_h$ and $\mathcal{X}_w$. Then, we transpose $\vb{Z_h}$ and $\vb{Z_w}$ to tensors of size $H\!\times\! W \!\times\! D$, say $\vb{Z_h}^T$ and $\vb{Z_w}^T$, respectively, where $\vb{Z_h}^T[i,j,k] = \vb{Z_h}[k,j,i]$ and $\vb{Z_w}^T[i,j,k] = \vb{Z_w}[i,k,j]$. Finally, we obtain the output feature tensor, $\vb{Z} \!\in\! \mathbb{R}^{H\!\times\! W \!\times\! D}$ by adding $\vb{Z_d}, \vb{Z_h}^T,$ and $\vb{Z_w}^T$. 
\vspace{-0.5em}
\begin{align}
    \vb{Z_d}[i,j,k] &= \vb{w_d}[k] ~~ \mathcal{X}[i,j,k] \\
    \vb{Z_h}[k,j,i] &= \vb{w_h}[i] ~~ \mathcal{X}_h[k,j,i]  \\
    \vb{Z_w}[i,k,j] &= \vb{w_w}[j] ~~ \mathcal{X}_w[i,k,j]  \\
    \vb{Z}[i,j,k] &= \vb{Z_d}[i,j,k] + \vb{Z_h}^T[i,j,k] + \vb{Z_w}^T[i,j,k]
\end{align}
\vspace{-1em}

\begin{figure}[t]
    \centering
    \begin{subfigure}[b]{0.32\linewidth}
		\centering
		\includegraphics[width=\linewidth,height=6em]{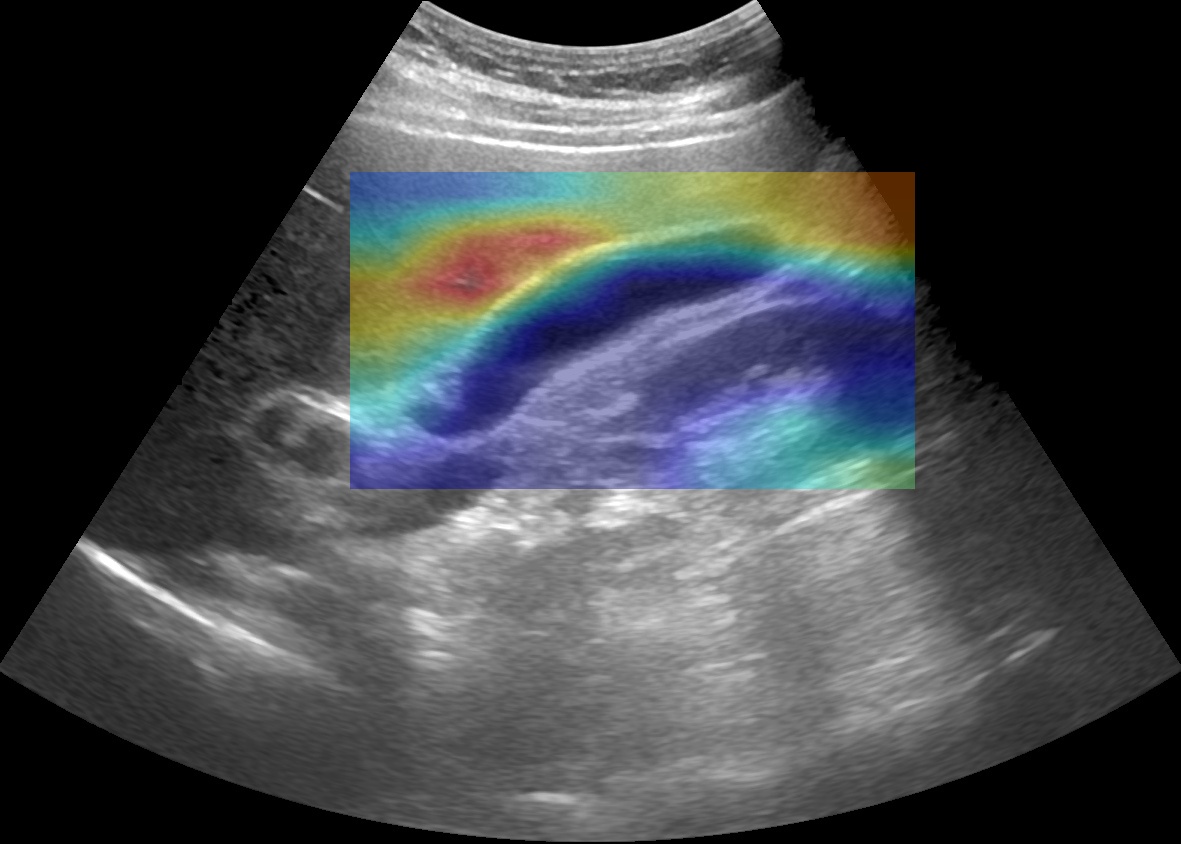}
		\caption{}
	\end{subfigure}
    \begin{subfigure}[b]{0.32\linewidth}
		\centering
		\includegraphics[width=\linewidth,height=6em]{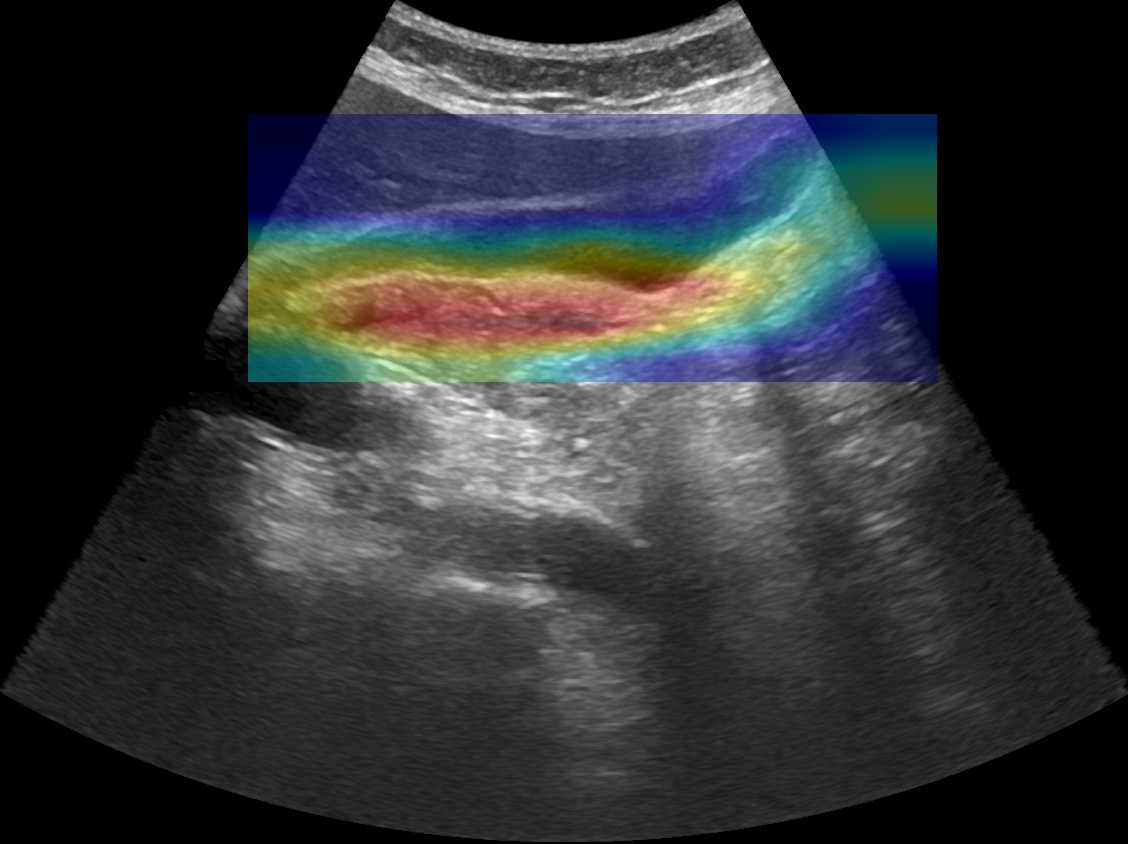}
		\caption{}
	\end{subfigure}
	\begin{subfigure}[b]{0.32\linewidth}
		\centering
		\includegraphics[width=\linewidth,height=6em]{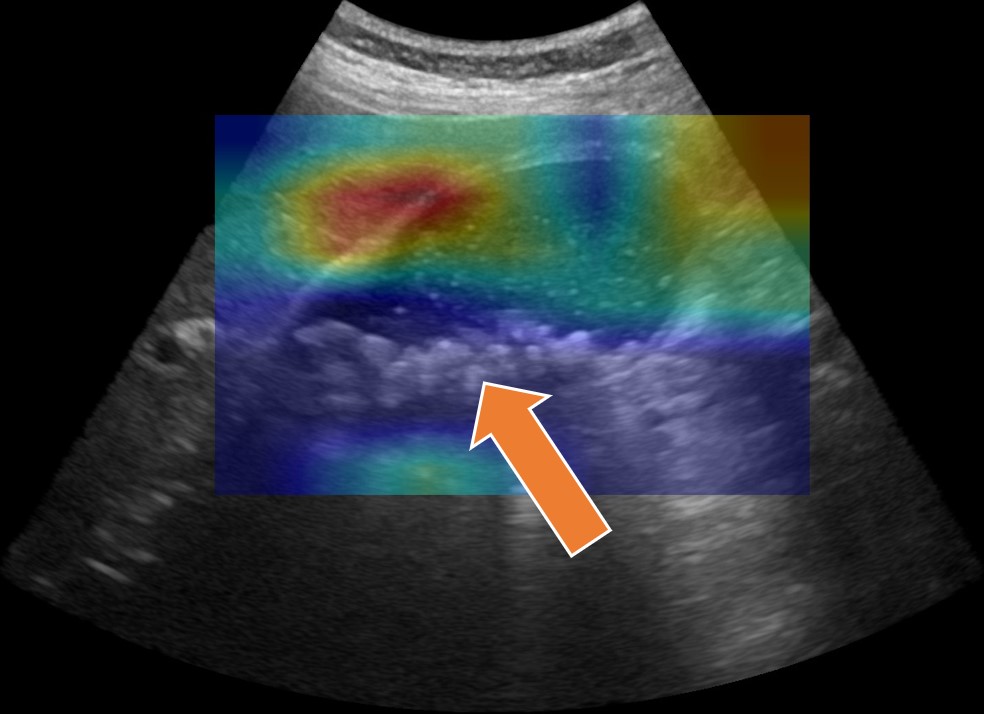}
		\caption{}
	\end{subfigure}
    \caption{Grad-CAM visual of GBCNet trained without curriculum showing how the model tends to get biased due to the presence of textures due to noise or organ tissue. GBCNet focuses on - (a) adjacent liver tissues than the normal \gb, (b) the echogenic region below the \gb, and (c) liver textures instead of the stones (highlighted using the arrow).}
    \label{fig:texture_bias_sample}
\end{figure}
%
\subsection{Visual Acuity Inspired Curriculum}
We found that the textures having visual characteristics of soft tissue can adversely affect the performance of GBCNet (\cref{fig:texture_bias_sample}). We propose a curriculum to mitigate the texture bias and improve the classification. We observed that while the MS-SoP classifier is affected by texture bias, the region selection network still maintains a very high recall (\cref{tbl:perf_region}). Hence, we used the curriculum training only on the classifier and not on the region selection network.

\mypara{Visual Acuity in Humans}
Visual Acuity (VA) refers to the clarity and sharpness of human vision. Due to the immaturity of the retina and visual cortex, newborn children have very low VA \cite{courage1990visual}. The VA improves with the maturation of the retina and visual cortex. However, for children with congenital cataracts, the cortex matures despite the lenticular opacity. Such children begin their visual activity with higher initial VA. Evidence shows that children with high initial VA suffer to facilitate spatial analysis over expansive areas \cite{vogelsang2018VisualAcuity}.  Low VA renders blurry images that do not contain enough local information for the visual cortex to identify patterns. As a result, the visual cortex tries to increase the receptive field to facilitate spatial analysis over expansive areas and learn global features \cite{kwon2016compensation, smith2009smile}. 

\begin{figure}[t]
	\centering
	\includegraphics[width=0.95\linewidth]{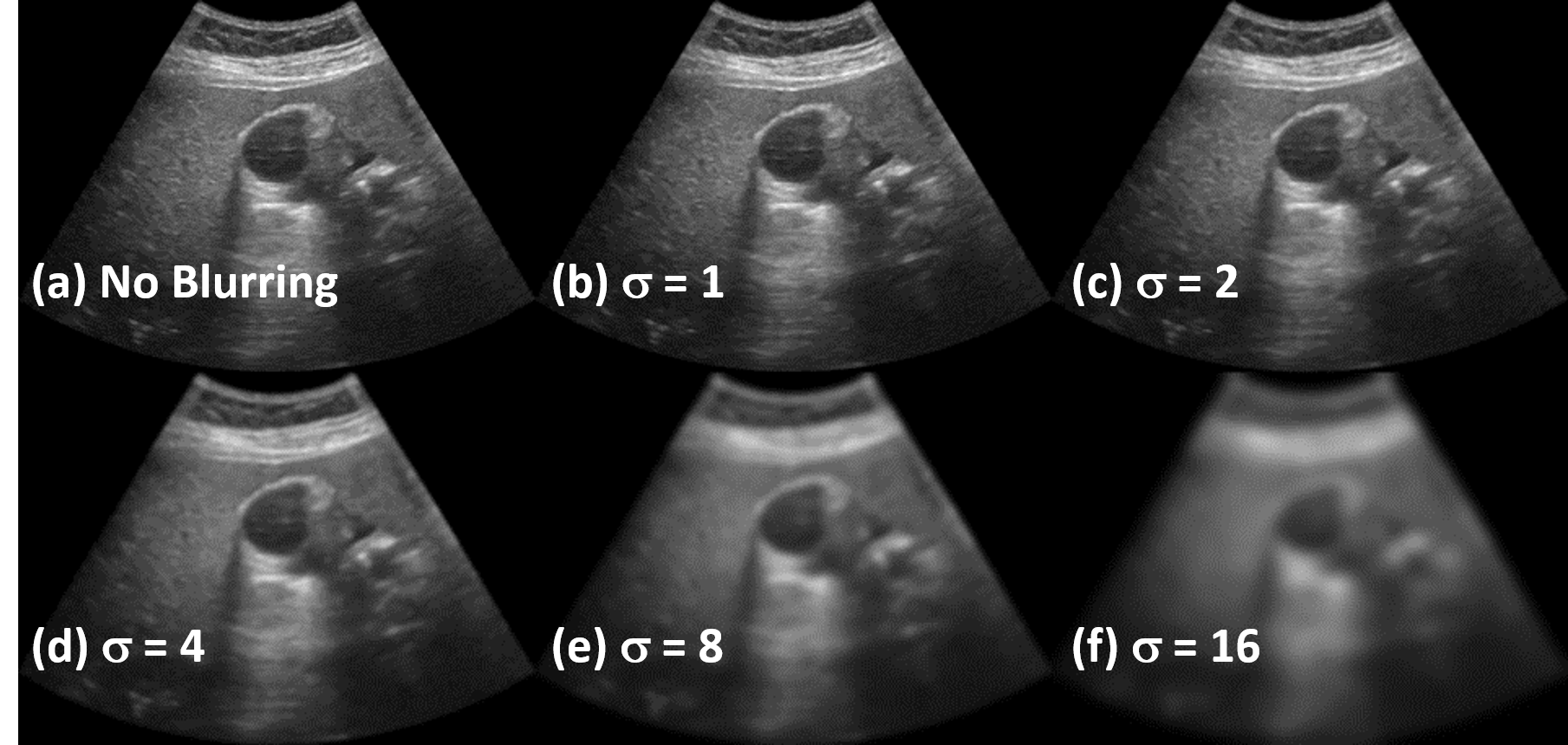}
	\caption{We simulate visual acuity through the Gaussian blur. Increasing $\sigma$ in a Gaussian filter decreases the visual acuity. Notice in the figure that, the effect of textures reduce as the visual acuity decreases and \gb shape and structure become more pronounced.}
	\label{fig:vis_acuity_sample}
\end{figure}

\mypara{Gaussian Blurring to Simulate Visual Acuity}
Gaussian filters are low-pass filters to cloak the high-frequency components of an input. A standard deviation $\sigma$ parameterizes the Gaussian filters. Increasing the $\sigma$ generates a higher amount of blur and low VA when convolved with an image. \cref{fig:vis_acuity_sample} shows how we can decrease the VA by increasing the $\sigma$ of a Gaussian filter. In our experiments, we have varied $\sigma$ from $1$ to $16$ to generate different levels of VA. 

\mypara{Proposed Curriculum}
While \cite{vogelsang2018VisualAcuity} demonstrates the improvement in receptive fields by gradually improving the sharpness of images during training, we take this observation further and show that the strategy of training on progressively higher resolution images also reduces the texture bias of a classification model. We propose a visual acuity-based training curriculum (\cref{algo}) that starts training the network with blurry and low-resolution \usg images and progressively increase the sharpness of training samples. The initial blurring allows the model to use an extended receptive field and focus on learning the global features such as the shape of the \gb while ignoring any noise or irrelevant textures. In the later phases, the sharp images allow the model to focus on the relevant local features in a controlled manner to make more accurate predictions. 

\begin{algorithm}[t]
\small
	\caption{Proposed VA-based curriculum}
	\label{algo}
	\SetAlgoLined
	\KwIn{$D^{\text{train}}$, Dataset of regions cropped from the original USG images.}
	\KwOut{Optimized model parameters $W^*$}
	Initialize $\sigma = \sigma_0$ \;
	Initialize model parameters to $W$ \;
	
	\For {epoch=$\{1 \ldots, N\}$}{
		\eIf{$\sigma > 0$}{
		    $Z = \phi$ \;
    		\For {$x \in D^{\text{train}}$} {
    			$Z = Z \cup \{x \oast G(\sigma)\}$ \;
    		}
		    train($W, Z$)\;
		}
		{
		    train($W, D^{\text{train}}$) \;
		}
		\If{(epoch $> k'$) and (epoch$\%k == 0$)}{
			$\sigma = \lfloor \sigma/2 \rfloor$ \; 
		}
	}
\end{algorithm}

\section{Dataset Collection and Curation}

\myfirstpara{Data Collection}
We acquired data samples from patients referred to PGIMER, Chandigarh (a tertiary care referral hospital in Northern India) for abdominal ultrasound examinations of suspected \gb pathologies. The study was approved by the Ethics Committee of PGIMER. We obtained informed written consent from the patients at the time of recruitment, and protect their privacy by fully anonymizing the data. Minimum 10 grayscale B-mode static images, including both sagittal and axial sections, were recorded by radiologists for each patient using a Logiq S8 machine. We excluded color Doppler, spectral Doppler, annotations, and measurements. Supplementary A 
contains more details of the data acquisition process.

\mypara{Labeling and ROI Annotation}
Each image is labeled as one of the three classes - normal, benign, or malignant. The ground-truth labels were biopsy-proven to assert the correctness. Additionally, in each image, expert radiologists have drawn an axis-aligned bounding box spanning the entire \gb and adjacent liver parenchyma to annotate the \roi. 

\mypara{Dataset Statistics}
We have annotated 1255 abdominal \usg images collected from 218 patients from the acquired image corpus. Overall, we have 432 normal, 558 benign, and 265 malignant images. Of the 218 patients, 71, 100, and 47 were from the normal, benign, and malignant classes, respectively. The width of the images was between 801 and 1556 pixels, and the height was between 564 and 947 pixels due to the cropping of patient-related information. 

\mypara{Dataset Splits}
The sizes of the training and testing sets are 1133 and 122, respectively. To ensure generalization to unseen patients, all images of any particular patient were either in the train or the test split. The number of normal, benign, and malignant samples in the train and test set is 401, 509, 223, and 31, 49, and 42, respectively. Additionally, we report the 10-fold cross-validation metrics on the entire dataset for key experiments to assess generalization. All images of any particular patient appeared either in the training or the validation split during the cross-validation. 

\begin{table}[t]
	\centering
	\resizebox{ \linewidth}{!}{%
		\begin{tabular}{lccccccc}
			\toprule[1pt]
			\multirow{2}{*}{\textbf{Method}} & \multicolumn{4}{c}{\textbf{Test Set}} & \multicolumn{3}{c}{\textbf{Cross Val.}} \\
			\cmidrule{2-8}
			& \textbf{Acc.} & \textbf{Acc.-2} & \textbf{Spec.} & \textbf{Sens.} & \textbf{Acc.} & \textbf{Spec.} & \textbf{Sens.}  \\
			\midrule[0.5pt]
			Radiologist A & 70.0 & 81.6 & 87.3 & 70.7 & -- & -- & --  \\
			Radiologist B & 68.3 & 78.4 & 81.1 & 73.2 & -- & -- & --  \\
			\midrule
			VGG16 & 62.3 & 72.1 & 90.0 & 38.1 & 69.3 $\pm$ 3.6 & 96.0 $\pm$ 4.6 & 49.5 $\pm$ 23.4 \\ 
			ResNet50 & 76.2 & 78.7 & 87.5 & 61.9 & 81.1 $\pm$ 3.1 &  92.6 $\pm$ 6.9 & 67.2 $\pm$ 14.7 \\ 
			InceptionV3 & 77.9 & 85.0 & 87.5 & 80.1 & 84.4 $\pm$ 3.9 & 95.3 $\pm$ 2.9 & 80.7 $\pm$ ~~9.7 \\
			Faster-RCNN & 71.3 & 77.9 & 76.2 & 81.0 & 75.7 $\pm$ 5.3 & 84.0 $\pm$ 4.6 & 80.8 $\pm$ 10.4 \\
			RetinaNet & 75.4 & 83.6 & 86.3 & 78.6 & 74.9 $\pm$ 7.3 & 86.7 $\pm$ 7.8 & 79.1 $\pm$ ~~8.9 \\
			EfficientDet & 58.2 & 77.9 & 86.3 & 62.0 & 73.9 $\pm$ 8.4 & 88.1 $\pm$ 9.9 & 85.8 $\pm$ ~~6.1 \\
			\midrule
			GBCNet & 87.7 & 91.0 & 90.0 & 92.9 & 88.2 $\pm$ 5.1 & 94.2 $\pm$ 3.7 & \textbf{92.3 $\pm$ ~~7.1} \\ 
			GBCNet+VA &\textbf{91.0} & \textbf{95.9} & \textbf{95.0} & \textbf{97.6} & \textbf{92.1 $\pm$ 2.9} & \textbf{96.7 $\pm$ 2.3} & 91.9 $\pm$ ~~6.3 \\
			\bottomrule[1pt]
		\end{tabular}
	}
	\caption{The model performances on the test set and the 10-fold cross validation (Mean$\pm$SD) in classifying \gbc from USG images. Apart from the standard accuracy of classifying normal, benign, and malignant \gb, we show the binary classification (malignancy vs. non-malignancy) accuracy on the test set (column Acc.-2). We also report the \gbc detection performance of two expert radiologists on the test set. The radiologists classified each test image without accessing the biopsy results or any other patient data. Note that our model significantly outperforms even the human radiologists. Recall that our ground truth labels are biopsy-proven. The performance of human radiologists in the our study is comparable to that reported in literature \cite{bo2019diagnostic, gupta2020evaluation}. }
	\label{tbl:perf_gbc}
\end{table}

\section{Implementation and Evaluation}

\myfirstpara{Transfer Learning and Data Augmentation}
Studies show that pre-training on natural image data improves network performance on medical image data \cite{alzubaidi2020transferlearning, cheng2017transfer}. We use \roi detection networks pre-trained on the COCO \cite{coco}, and classifiers pre-trained on ImageNet data \cite{imagenet}. We use resizing, center-cropping, and normalization data augmentations to avoid over-fitting on the small \usg dataset.

\mypara{Hyper-parameters}
Details of the hyper-parameters are in the supplementary C. 
We train and freeze the \roi detection network before training the classifier. We used $\sigma_0\!=\!16$, $k'\!=\!10$, $k\!=\!5$, for the curriculum.

\mypara{Evaluation metrics}
We use mean intersection over union (mIoU), precision, and recall for comparing region selection models. We compute the precision and recall as suggested by \cite{ribli2018detecting} (details in the supplementary D). 
To assess the classification models for \gbc detection, we use accuracy, sensitivity, and specificity as the evaluation metrics.

\begin{figure}[t]
	\centering
	\includegraphics[width=\linewidth]{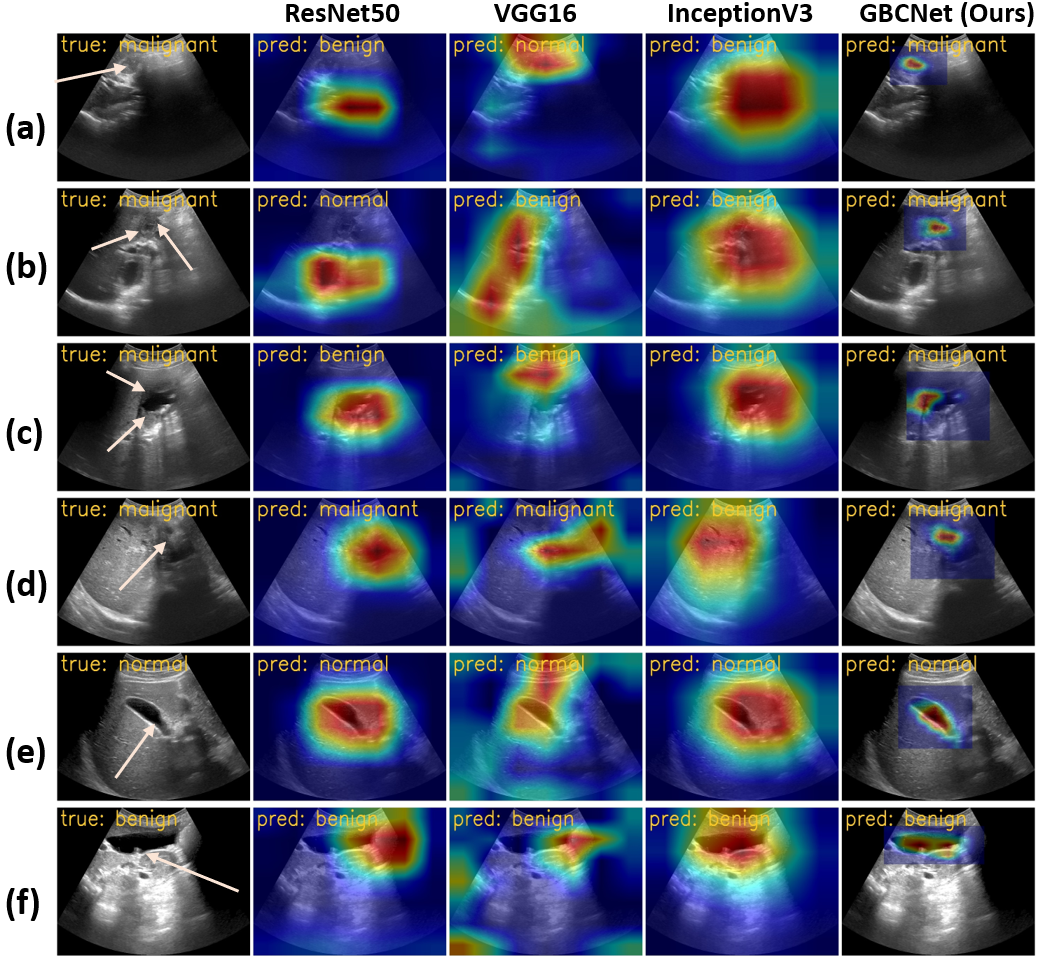}
	\caption{Grad-CAM visuals and the predictions for ResNet50, VGG16, Inception-V3, and GBCNet. The pathological areas are shown with arrows in the original images. 
	(a) ResNet50 and Inception-V3 focus on the shadow, whereas VGG16 focuses on the echogenic area, and all three fail to detect \gbc. GBCNet accurately focuses on the malignant \gb region invading the liver and detects \gbc.  (b), (c) The baseline networks focus on shadow or noise instead of the cancerous area and mispredict. 
	(d) Although ResNet50 and VGG16 predict malignancy, they fail to precisely focus on the malignant region compared to GBCNet. Inception-V3 failed to classify \gbc. (e), (f) GBCNet pinpoints the discriminating region compared to the baselines for normal and benign \gb regions, respectively. More visuals provided in supplementary E.} 
	\label{fig:gbc_vis}
\end{figure}

\section{Experiments and Results}
\subsection{Efficacy of GBCNet over Baselines}
We compare GBCNet with three popular deep classifiers, ResNet-50 \cite{resnet}, VGG-16 \cite{vgg}, and Inception-V3 \cite{inception}. We also evaluate the performance of three \sota object detectors, Faster-RCNN \cite{fasterrcnn}, RetinaNet \cite{retinanet}, and EfficientDet \cite{efficientdet} for detecting \gbc. We report the results in \cref{tbl:perf_gbc}. From the reported results, it is clear that baseline networks have poor accuracy for detecting \gbc from \usg images. Grad-CAM \cite{gradcam} visualizations in \cref{fig:gbc_vis} show that the noise, textures, and artifacts significantly influence the decision of baseline classification models. As a result, their classification accuracy suffers heavily. Compared to the baselines, GBCNet along-with the proposed MS-SoP classifier precisely focuses on crucial visual cues leading to its superior performance.

\begin{table}[t]
	\centering
	\scriptsize
	\resizebox{\linewidth}{!}{%
    \begin{tabular}{@{}lccc@{}}
    \toprule[1pt]
    \textbf{Model} & \textbf{mIoU} & \textbf{Precision} & \textbf{Recall} \\
    \midrule[0.5pt]
    Faster-RCNN & 71.1 $\pm$ 2.7 &  96.0 $\pm$ 2.6 & 99.2 $\pm$ 0.7\\
    YOLOv4  & 70.7 $\pm$ 2.9 & 98.1 $\pm$ 2.3 & 97.9 $\pm$ 1.5 \\
    CentripetalNet & 60.4 $\pm$ 4.7 & 95.1 $\pm$ 3.8 & 89.6 $\pm$ 7.3 \\
    Reppoints & 69.1 $\pm$ 3.2 & 95.2 $\pm$ 3.9 & 99.7 $\pm$ 0.4\\
    \bottomrule[1pt]
    \end{tabular}
	}
	\caption{Comparison of the \gb region selection models. We reported 10-fold cross validation (Mean$\pm$SD) of the metrics.}
\label{tbl:perf_region}
\end{table}
\begin{figure}[t]
	\centering
	\includegraphics[width = \linewidth]{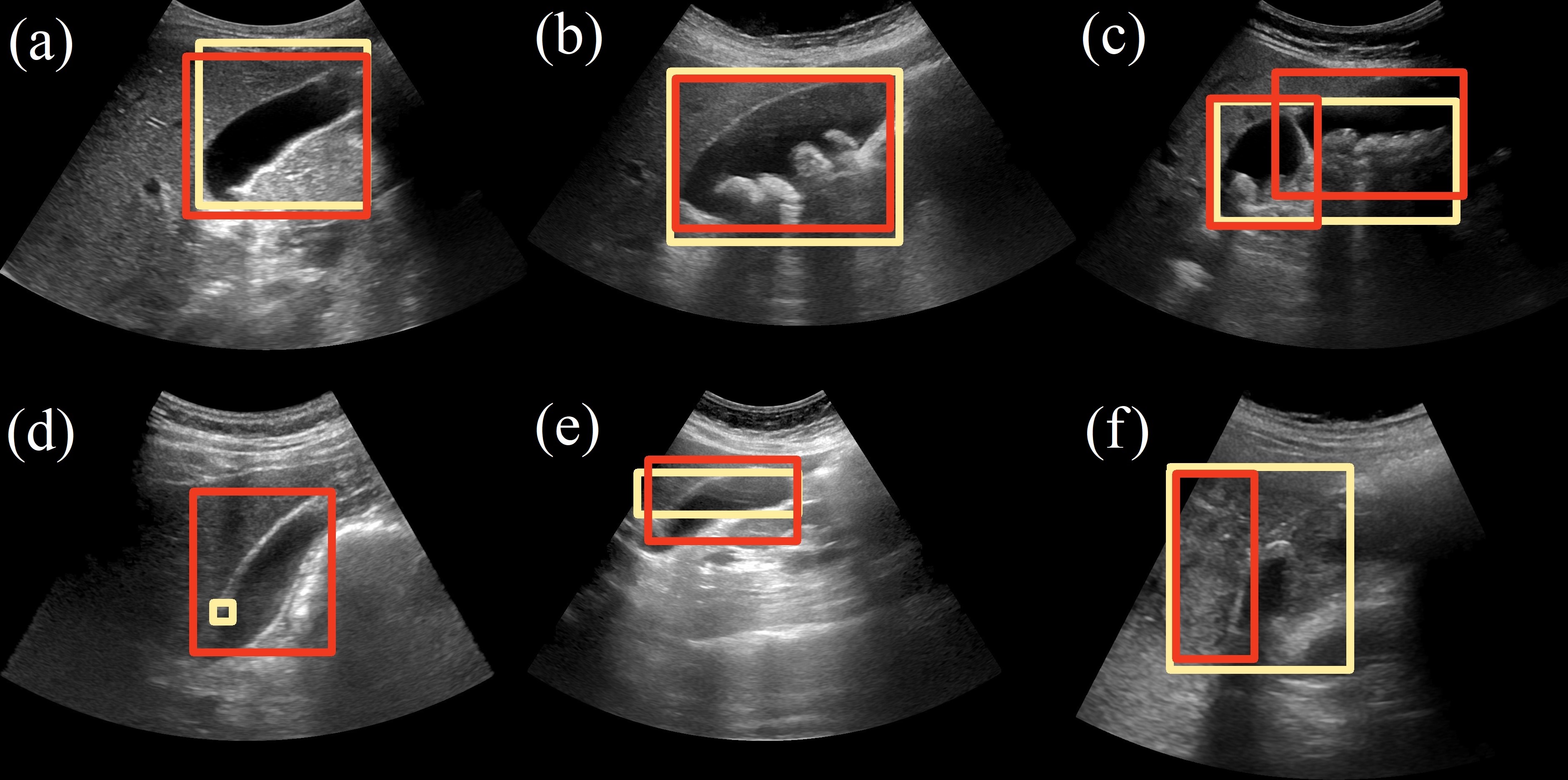}
	\caption{We visually compare \roi selection by Faster-RCNN (dark red) with the \roi identified by expert Radiologists (light yellow). (a, b) The predicted a\roi matches well with the radiologists' expectations. (c) The model considers the sections partitioned by the \gb wall as separate regions. However, the union of the predicted boxes very closely approximates the actual \gb region. (d, e) Although the radiologist made an error in judging the \roi, Faster-RCNN was able to identify an accurate \roi resulting in a visually superior prediction. (f) The predicted \roi covers only a portion of the area an expert radiologist considered necessary. Even though the region prediction seems inferior compared to the human perception, expert radiologists corroborated that the predicted region captures the \gb invading the liver, a vital visual cue to detect \gbc. \roi samples from other detectors are in supplementary F.}
	\label{fig:region_vis}
\end{figure}
%
%
\subsection{Performance of GB Region Selection Models}
\cref{tbl:perf_region} summarizes the performance of various models for localizing the \gb region. For critical tasks such as region selection for cancer detection, recall is more important than precision. Multiple predicted regions can be discarded in the second stage, but missing any potentially malignant region could be disastrous. We note that the Faster-RCNN achieves the highest mIoU out of all the models while maintaining very high recall and excellent precision. Hence, we use Faster-RCNN as the region selection model. In \cref{fig:region_vis} we show the visual comparison of the \gb localization results of Faster-RCNN along with the \rois annotated by the expert radiologists. The model could predict the region of interest accurately in most cases. Although the model's prediction visually differed from the radiologists in some samples, closer inspection revealed that the predicted region retains sufficient visual cues to detect malignancy. 

\begin{table}[t]
	\centering
    \begin{tabular}{@{}lccc@{}}
    \toprule[1pt]
    \textbf{Model} & \textbf{Spec.} & \textbf{Sens.} \\
    \midrule[0.5pt]
    ResNet50 &  97.5 $\pm$ 2.4 & 82.9 $\pm$ 8.8\\
    DenseNet121  & 96.8 $\pm$ 1.8 & 82.4 $\pm$ 2.7 \\
    \midrule[0.5pt]
    MS-SoP (ours) & 96.7 $\pm$ 2.7 & 87.1 $\pm$ 7.1 \\
    \bottomrule[1pt]
    \end{tabular}
	\caption{The sensitivity and specificity of MS-SoP and two baseline classifiers on breast cancer detection from USG images. We report 5-fold cross-validation on the BUSI dataset.}
\label{tbl:busi}
\end{table}

\begin{table}[t]
	\centering
    \resizebox{ \linewidth}{!}{%
    \begin{tabular}{lcccc}
    \toprule[1pt]
    \multirow{2}{*}{\textbf{Model}} & \multicolumn{2}{c}{\textbf{Orig. Test Set}} & \multicolumn{2}{c}{\textbf{Synth. Test Set}} \\
    & \textbf{Spec.} & \textbf{Sens.} & \textbf{Spec.} & \textbf{Sens.}\\
    \midrule[0.5pt]
    ROI+VGG16 & 83.8 & 57.2 & 78.7 ($\downarrow$~~6.1) & 57.2 \\
    ROI+VGG16+VA & 82.5 & 76.2 & 77.5 ($\downarrow$~~6.1) & 76.2 \\
    \midrule
    ROI+ResNet50 & 86.3 & 85.7 & 65.0 ($\downarrow$24.7)& 85.7 \\
    ROI+ResNet50+VA & 93.8 & 85.7 & 88.7 ($\downarrow$~~5.4) & 85.7 \\
    \midrule
    ROI+Inception-V3 & 56.3 & 83.3 & 41.3 ($\downarrow$26.6) & 83.3 \\
    ROI+Inception-V3+VA & 91.3 & 69.0 & 78.8 ($\downarrow$13.7) & 69.0 \\
    \midrule
    GBCNet & 90.0 & 92.9 & 76.2 ($\downarrow$15.3) & 92.9 \\
    GBCNet+VA & 95.0 & 97.6 & 85.0 ($\downarrow$10.5) & 97.6 \\
    \bottomrule[1pt]
    \end{tabular}
    }
    \caption{Robustness of the curriculum in tacking texture bias while detecting \gbc. We show the performance of using curriculum on four models that apply classifiers on localized \gb region - (a) ROI+VGG16, (b) ROI+ResNet50, (c) ROI+Inception-V3, and (d) GBCNet (ROI+MS-SoP). The relative change (in percentage) in specificity for synthetic test data is shown within parentheses. The sensitivity remains unchanged as the malignant images were not altered. Observe that as compared to the models trained on high-resolution images, our VA-based curriculum is more robust to textures and is able to maintain a lower drop in specificity. The only exception is the ROI+VGG16 model, for which the curriculum training does not lower the drop in specificity. }
\label{tbl:curr_texture}
\end{table}

\subsection{Applicability of the Proposed Classifier in Breast Cancer Detection from USG Images}
We explored the applicability of the proposed MS-SoP classifier on breast cancer detection from \usg images for a publicly available dataset, BUSI \cite{al2020dataset}, containing 133 normal, 487 benign, and 210 malignant images. The images in BUSI are already cropped from original \usg images to highlight only the important regions. Thus, we skip the \roi selection and run the MS-SoP classifier on BUSI. \cref{tbl:busi} shows that the MS-SoP classifier achieves much better sensitivity, which indicates the superiority of the MS-SoP architecture for malignancy identification on \usg images. We note that while breast cancer detection relies on tissue/ mass characterization, \gbc detection is primarily based on wall shape and mass anomaly. 

\subsection{Efficacy of the Proposed Curriculum}
\begin{figure}[t]
	\centering
	\includegraphics[width=0.95\linewidth]{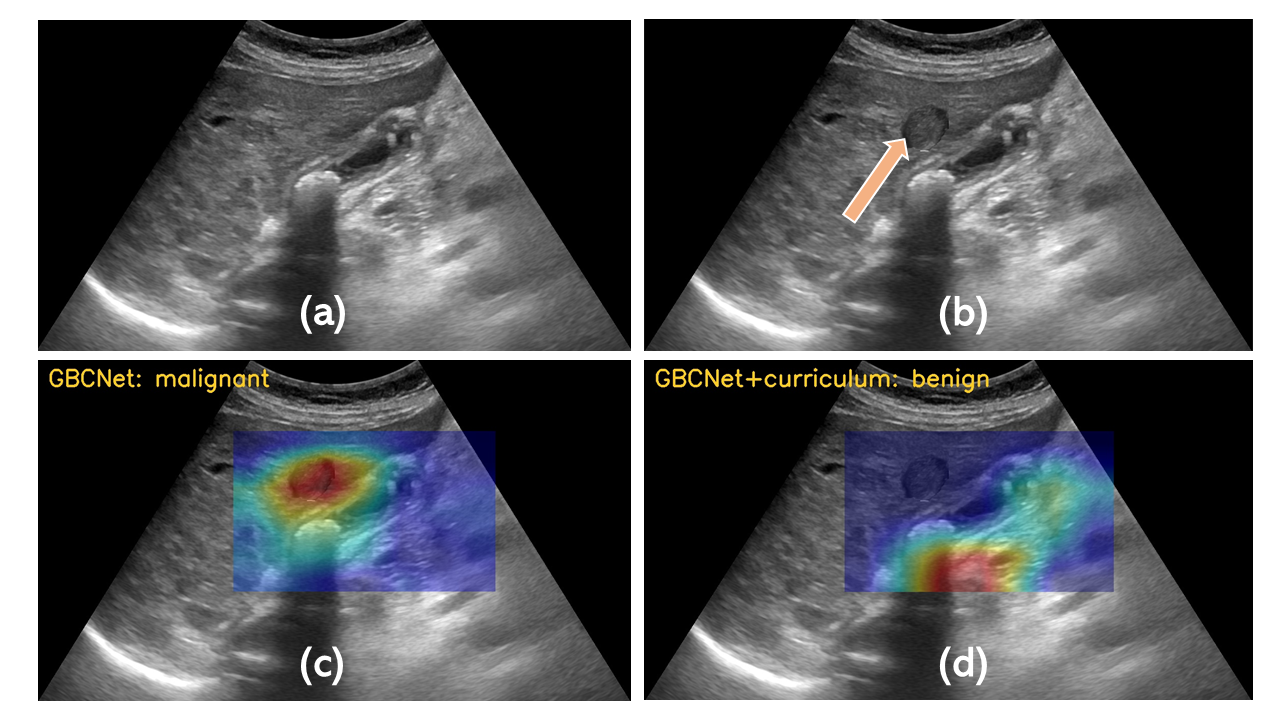}
	\caption{(a) Original image of a benign \gb. The \gb presents a stone and thickened wall. (b) In the synthetic image, we added an artificial tissue-like patch near the benign \gb region (highlighted by the arrow). This patch is not a part of the original \gb, and expert radiologists confirmed that the diagnosis of the \gb is not altered. (b) The textured artificial patch makes the GBCNet biased, and it focuses on the patch to predict the sample as malignant (false-positive). (d) Visual acuity curriculum fixes the texture bias of the GBCNet and helps the network to re-adjust the salient regions to the actual \gb pathology. 
}
	\label{fig:texture_vis}
\end{figure}

\myfirstpara{Robustness in Tackling Texture Bias}
As described earlier, spurious textures present in \usg images tend to increase the false positives in detecting malignancy. To validate this hypothesis, we created a synthetic test set. We used the method given by \cite{yang2020fda} and added low-level frequencies from malignant images to alter the original texture of normal and benign samples. We also manually added patches looking like soft tissue near the \gb region of normal and benign images. Expert radiologists confirmed that the diagnosis of the \gb pathology is not altered. \cref{tbl:curr_texture} shows that the specificity decreases due to the increase of the false positives. The sensitivity remains unaffected as the prediction for malignant \gb samples was unchanged. The network trained with the proposed curriculum tackles texture bias well and accurately predicts non-malignant samples in synthetic test data. This experiment shows that the proposed curriculum effectively tackles texture bias. \cref{fig:texture_vis} shows a visual sample of how the soft tissue-like texture influences the network's decision and how the curriculum helps the network to rectify the discriminative regions. 

\mypara{Performance Improvement} 
We also assess the quantitative performance improvement of models due to the curriculum training in supplementary B. 

\subsection{Ablation Study}
\begin{figure}[t]
	\centering
	\begin{subfigure}[b]{0.49\linewidth}
    \includegraphics[width=\linewidth]{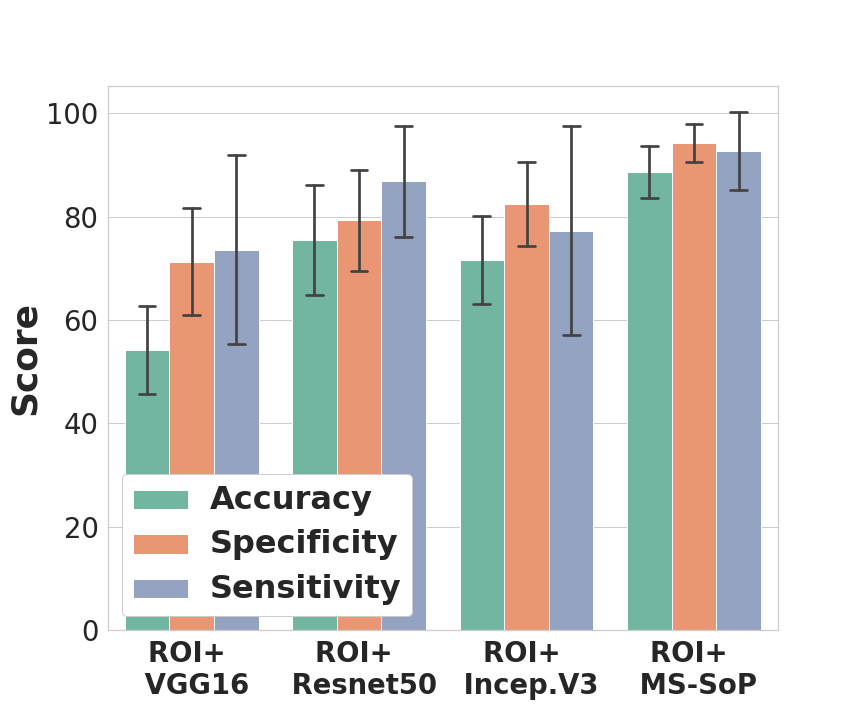}
    \caption{}
    \label{fig:perf_attn_models}
    \end{subfigure}
	\begin{subfigure}[b]{0.49\linewidth}
		\centering
		\includegraphics[width=\linewidth]{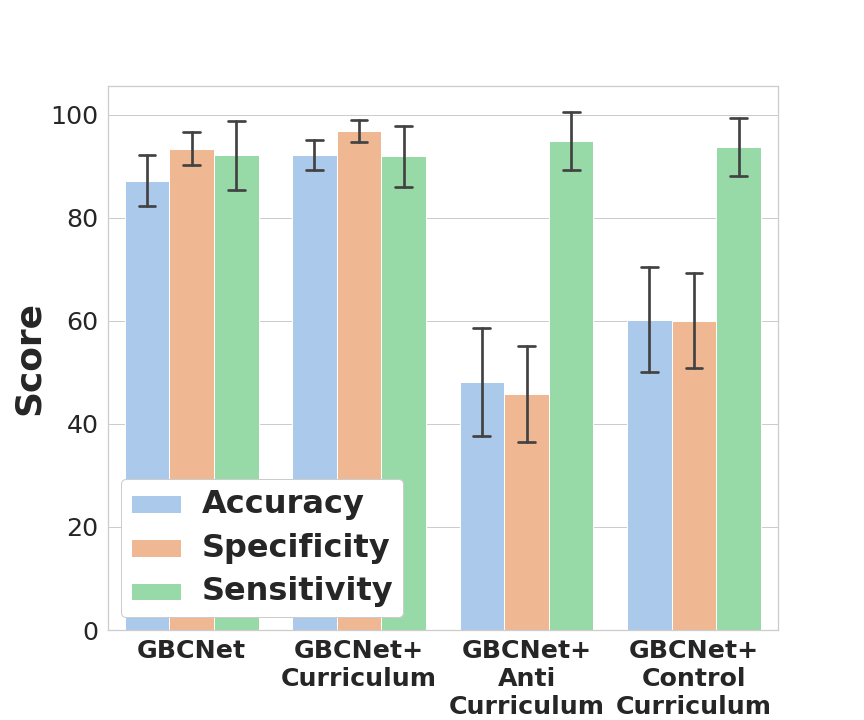}
		\caption{}
		\label{fig:ablation1}
	\end{subfigure}
	
	\begin{subfigure}[b]{0.49\linewidth}
		\centering
		\includegraphics[width=\linewidth]{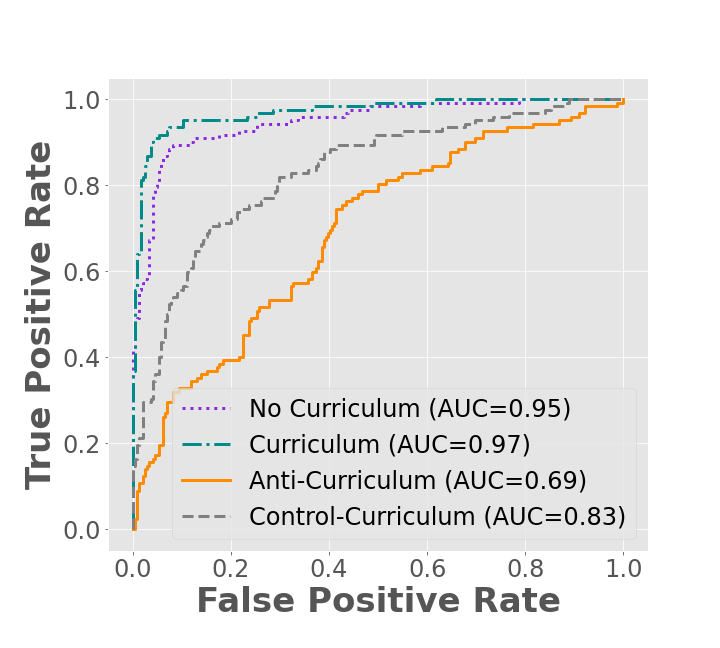}
		\caption{}
		\label{fig:ablation2}
	\end{subfigure}
		\begin{subfigure}[b]{0.49\linewidth}
		\centering
		\includegraphics[width=\linewidth]{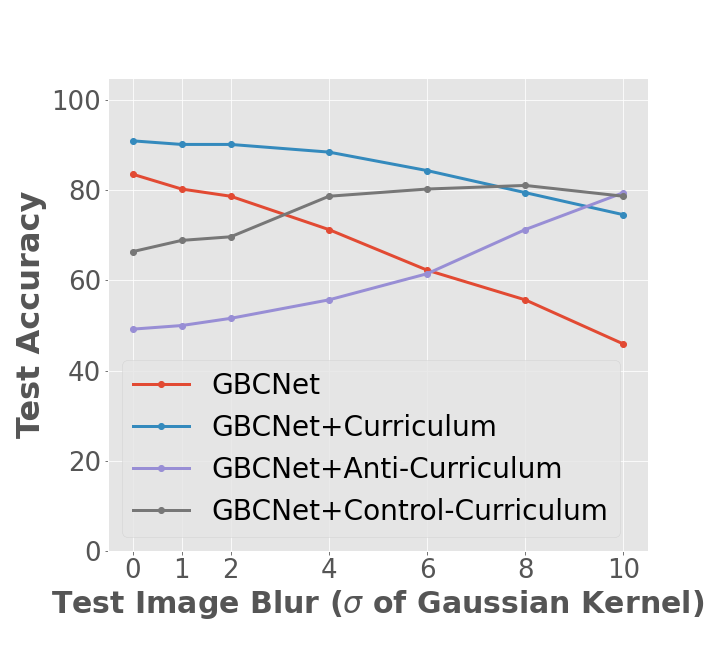}
		\caption{}
		\label{fig:ablation3}
	\end{subfigure}
	\caption{Ablation study. (a) Comparison of accuracy, specificity, and sensitivity for applying different classification networks (VGG16, ResNet50, Inception-V3, and MS-SoP) on the localized \gb. We have reported the 10-fold cross validation results. (b) The efficacy of the proposed training regime in terms of accuracy, specificity, sensitivity. (c) ROC-AUC for different training regimes on the test set. (d) The proposed curriculum generalizes better at different resolutions. }
	\label{fig:diff_images}
\end{figure}

\myfirstpara{Choice of Classifier in GBCNet}
We have plugged in other deep classification networks in place of the proposed MS-SoP classifier in the GBCNet framework. \cref{fig:perf_attn_models} summarizes the results. The MS-SoP classifier on GBCNet provides the best \gbc detection accuracy. Using the classifiers on the \rois improves the sensitivity and accuracy for ResNet50 and VGG16. However, the drop in specificity results in performance degradation for Inception-V3 as the sensitivity was not improved.  

\mypara{Choice of Training Regime}
We used the proposed GBCNet model to assess the influence of the visual acuity-based curriculum. We compare the curriculum with two possible alternatives - (i) \emph{anti-curriculum} that initially trains with high resolution and progressively lowers the resolution, and (ii) \emph{control-curriculum} where the samples are not sorted resolution-wise, and the curriculum contains a random set of blurred samples. Note that the control-curriculum can also be thought of using Gaussian blurring as a data augmentation where the probability of choosing a particular $\sigma$ is equal to the fraction of epochs the $\sigma$ used during the curriculum. \cref{fig:ablation1} and \ref{fig:ablation2} show the performance of the various curriculum strategies on the performance on GBCNet.  
Further, to understand how various curriculum strategies affect a model's generalization at different resolutions, we blur the test set using different values of $\sigma$ and evaluate the models on these images (\cref{fig:ablation3}). 
We see that the model trained using the proposed curriculum generalizes well across different image resolutions, which is an indicator of better spatial understanding.

\section{Conclusion}
This paper addresses Gallbladder Cancer detection from Ultrasound images using deep learning and proposes a new supervised learning framework (GBCNet) based on \roi selection and multi-scale second-order pooling. The proposed design helps the classifier focus on the crucial \gb region predicted by the region selection network. We propose a visual acuity-based curriculum to make our design resilient to texture bias and improve its specificity. Extensive experiments show that GBCNet, combined with curriculum learning, improves performance over the baseline deep classification and object detection architectures. We hope our work will generate interest in the community towards this important but hitherto overlooked problem of \gbc detection.

{\small \mypara{Acknowledgement} The authors thank IIT Delhi HPC facility for computational resources.}

\clearpage
{\small
\bibliographystyle{ieee_fullname}
\bibliography{main}
}

\clearpage
\section*{Supplementary Material}

\maketitle

\appendix
\beginsupplement

\section{Details of Data Acquisition and Annotation}
\label{supp:data_collection}
\myfirstpara{Data Acquisition} The study was approved by the ethics committee of the Postgraduate Institute of Medical Education and Research, Chandigarh. We performed all procedures according to the Declaration of Helsinki and the research guidelines of Indian Council of Medical Research. According to the hospital's protocol, 6 hours fasting was advised a day before the Ultrasound (USG) examinations for adequate distension of the GB. Two radiologists with expertise in abdominal USG performed the examinations on a Logic S8 machine (GE Healthcare) using a convex low-frequency transducer with a frequency range of 1--5 MHz.  USG assessment was done from different angles using both subcostal and intercostal views to visualize the entire GB, including the fundus, body, and neck.  Patients were examined in different positions for adequate visualization of the GB.  The screen area was adjusted so that the GB could occupy at least 20\% of the entire screen.  

\begin{figure}[h]
    \centering
     \begin{subfigure}[b]{0.32\linewidth}
		\centering
		\includegraphics[width=\linewidth, height=7em]{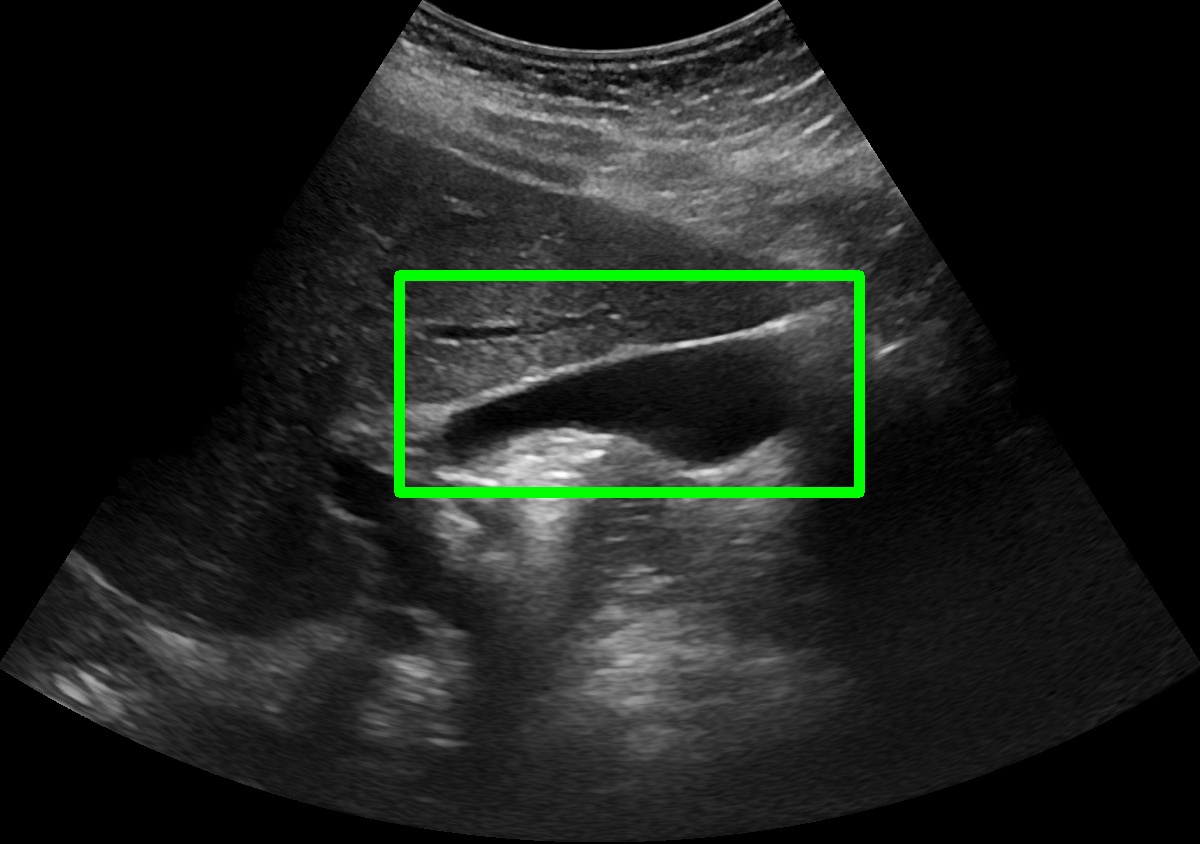}
		\caption{}
	\end{subfigure}
    \begin{subfigure}[b]{0.32\linewidth}
		\centering
		\includegraphics[width=\linewidth, height=7em]{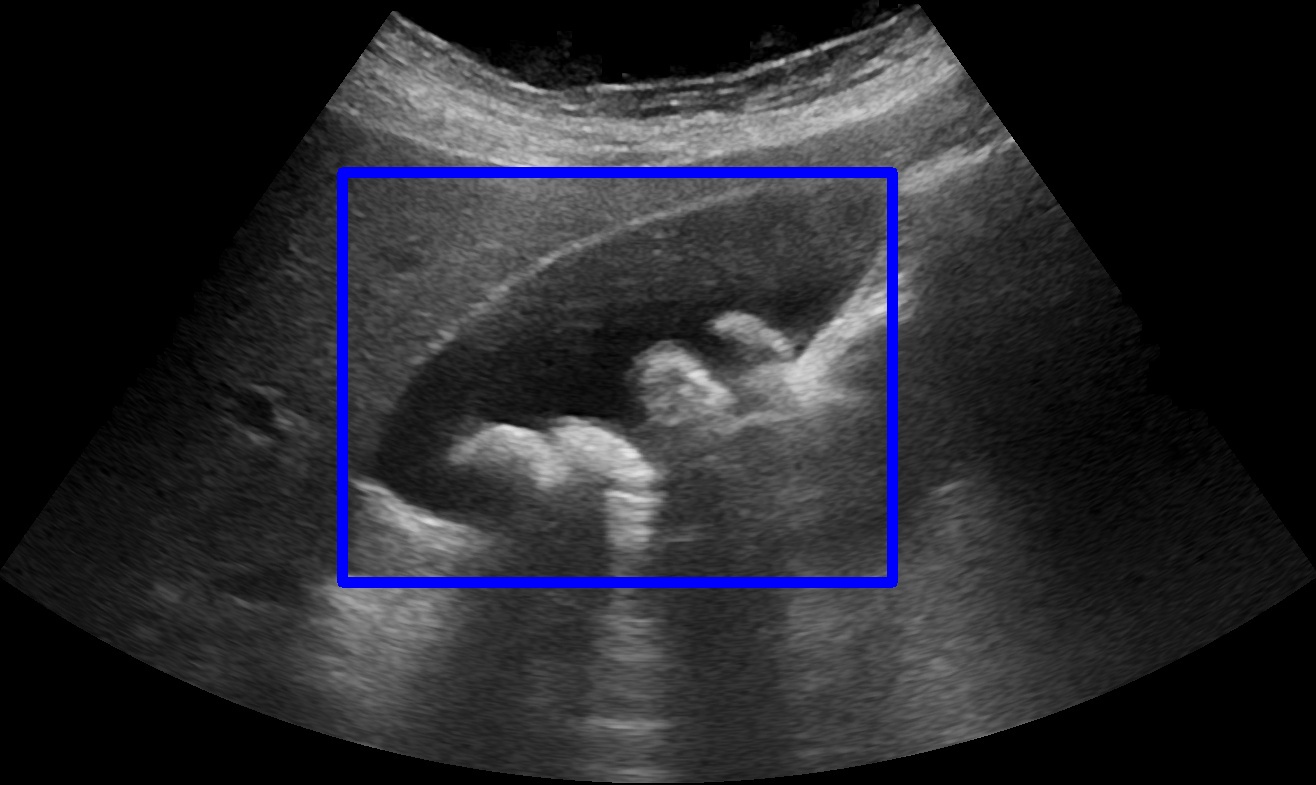}
		\caption{}
	\end{subfigure}
    \begin{subfigure}[b]{0.32\linewidth}
		\centering
		\includegraphics[width=\linewidth, height=7em]{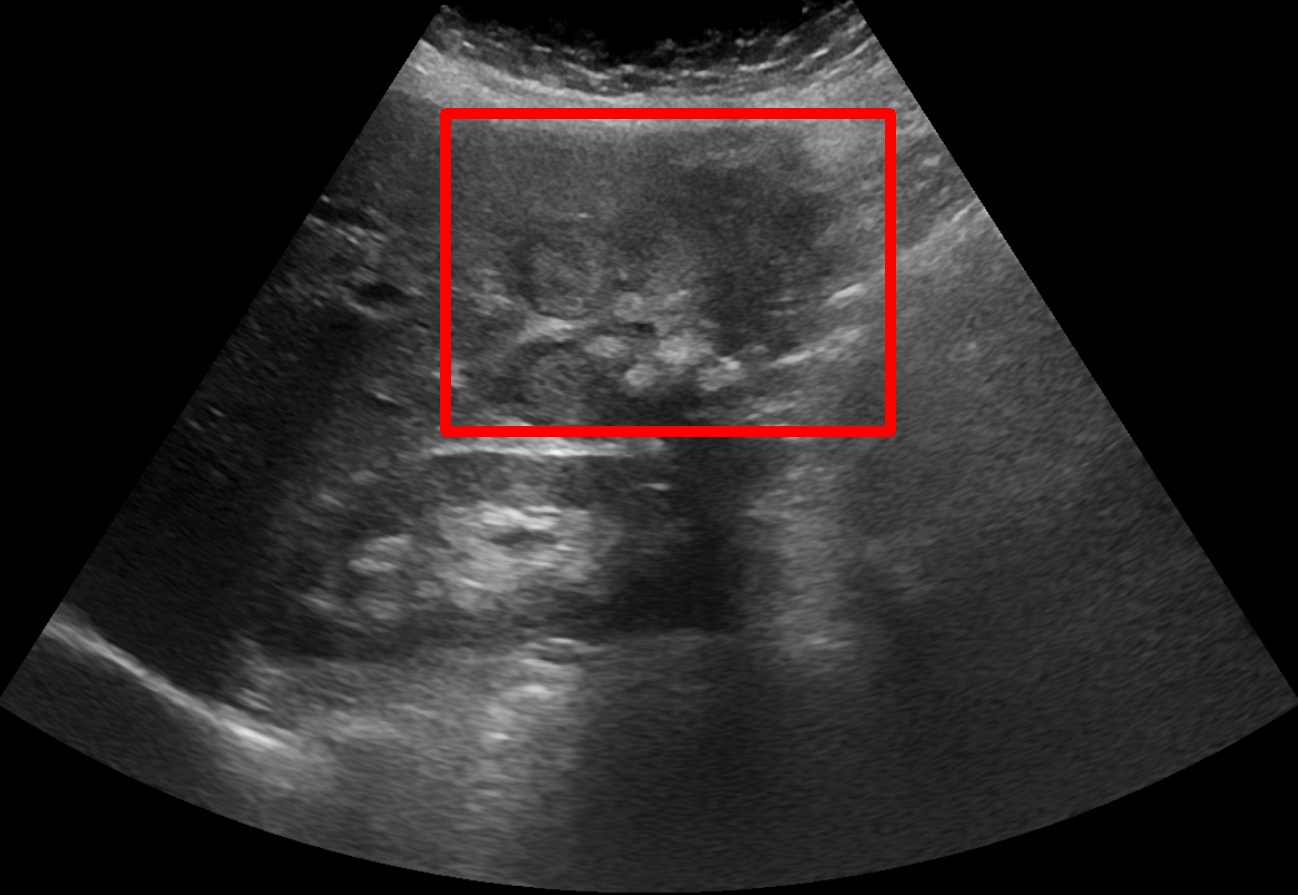}
		\caption{}
	\end{subfigure}
   \caption{Sample ROI annotation. (a) Normal GB with ROI annotated in green, (b) GB with benign abnormalities with ROI in blue, and (c) Malignant GB with ROI annotated in red. }
    \label{fig:bb_sample}
\end{figure}
\myfirstpara{ROI Annotation} Apart from image classification labels, we used bounding-box annotations to capture the GB localization. Two radiologists with 7 and 2 years of experience in abdomen radiology did the bounding-box annotations with consensus using the LabelMe \cite{russell2008labelme} software. A single free-size axis-aligned rectangular box in every image, spanning the entire GB and adjacent liver parenchyma, preferably keeping the GB in the box's center, highlights the region of interest (see \cref{fig:bb_sample}).

\begin{table}[t]
	\centering
	\resizebox{\linewidth}{!}{%
    \begin{tabular}{@{}lccc@{}}
    \toprule[1pt]
    \textbf{Model} & \textbf{Acc} & \textbf{Spec.} & \textbf{Sens.} \\
    \midrule[0.5pt]
    ROI+VGG16 & 53.3 $\pm$ 9.2 & 71.9 $\pm$ 11.5 & 73.3 $\pm$ 17.9\\
    ROI+VGG16+VA & 77.7 $\pm$ 4.1 & 93.8 $\pm$ 3.0 & 72.0 $\pm$ 19.5\\
    \midrule
    ROI+ResNet50 & 76.6 $\pm$ 10.7 &  82.3 $\pm$ 10.5 & 90.9 $\pm$ 11.1\\
    ROI+ResNet50+VA & 85.4 $\pm$ 7.7 & 92.3 $\pm$ 5.9 & 87.5 $\pm$ 9.1 \\
    \midrule
    ROI+Inception-V3 & 71.8 $\pm$ 8.9 & 83.3 $\pm$ 8.7 & 78.5 $\pm$ 21.4\\
    ROI+Inception-V3+VA & 82.6 $\pm$ 4.6 & 93.1 $\pm$ 4.4 & 82.6 $\pm$ 9.9\\
    \midrule
    RetinaNet & 74.9 $\pm$ 7.3 &  86.7 $\pm$ 7.8 & 79.1 $\pm$ 8.9\\
    RetinaNet+VA & 73.3 $\pm$ 6.0 & 92.1 $\pm$ 4.4 & 70.6 $\pm$ 14.2\\
    \midrule
    GBCNet (ROI+MS-SoP) & 88.2 $\pm$ 5.1 & 94.2 $\pm$ 3.7 & 92.3 $\pm$ 7.1\\
    GBCNet+VA & 92.1 $\pm$ 2.9 &  96.7 $\pm$ 2.3 & 91.9 $\pm$ 6.3\\
    \bottomrule[1pt]
    \end{tabular}
	}
	\caption{Model performances (10-fold cross-validation) for training with our proposed visual acuity-based curriculum.}
\label{tbl:curr_improve}
\end{table}

\section{Performance Improvement with Proposed Curriculum}
\label{supp:curr_improve}
We show the performance improvement of various models with the curriculum-based training in \cref{tbl:curr_improve}. All models show improvement in specificity, which indicates the effectiveness of the proposed blurring-based curriculum in tackling texture bias. 

\section{Implementation details}
\label{supp:impl}
\cref{tbl:configs} lists the configurations of all models which we have used. We trained on the Quadro P5000 16GB GPU. The table includes a brief description of the various stages of the network, input image sizes ($H\times W\times D$), the optimizer, relevant hyper-parameters such as learning rate, weight decay, momentum, batch size, and the number of training epochs/steps for the network. 

\begin{table*}[ht]
\centering
\resizebox{ \linewidth}{!}{%
\begin{tabular}{p{0.12\linewidth}p{0.4\linewidth}p{0.06\linewidth}p{0.2\linewidth}p{0.05\linewidth}p{0.06\linewidth}}
\toprule[1.5pt]
\textbf{Model} & \textbf{Description} & \textbf{Input Size} & \textbf{Optimizer} & \textbf{Batch size} & \textbf{Epochs/ Steps}
\\ \midrule[0.75pt]
YOLOv4 \cite{yolov4} & CSPDarknet53 backbone, PANet neck, anchor-based YOLO head. Total 162-layers. Backbone was frozen for first 800 step. Entire network was trainable thereafter. Single stage, anchor-based  & $608\times608\times3$ & SGD LR = 0.0001 momentum = 0.95 weight decay = 0.0005 & 64 & 3000 steps \\ \hline
Faster-RCNN \cite{fasterrcnn} & Resnet50 Feature Pyramid backbone. Backbone was frozen for training. Two-stage, anchor-based. & $800 \times 1333 \times 3$ & SGD LR = 0.005 momentum = 0.9 weight decay = 0.0005 & 16 & 60 epochs \\ \hline
Reppoints \cite{reppoints} & Resnet101 backbone, Group Normalization neck, and a reppoints head. Backbone was frozen for first 30 epochs, and entire network was trainable thereafter. Two-stage, anchor-free & $800 \times 1333 \times 3$ & SGD LR = 0.001 momentum = 0.9 weight decay = 0.0001 & 4 & 50 epochs \\ \hline
Centripetal-Net \cite{centripetalnet} & Improvement over CornerNet model. Uses centripetal shift to match corners. HourglassNet-104 backbone. Enitre network was trainable. Anchor-free & $511 \times 511 \times 3$ & Adam LR = 0.0005 & 4 & 50 epochs \\ \hline
ResNet \cite{resnet} & Resnet-50 used. All layers were trainable. Output dimension of last fully connected layer is three - corresponding to normal, benign, and malignant GB. LR decays by 10\% after every 5 epochs through a step LR scheduler. & $224\times224\times3$ & SGD LR = 0.005 momentum = 0.9 weight decay = 0.0005 & 16 & 100 epochs \\ \hline
VGG \cite{vgg} & VGG-16 is used. All layers were trainable. LR decays by 10\% after every 5 epochs through a step LR scheduler. & $224\times224\times3$ & SGD LR = 0.005 momentum = 0.9 weight decay = 0.0005 & 16 & 100 epochs \\ \hline
Inception \cite{inception} & Inception-V3 used. All layers were trainable. LR decays by 10\% after every 5 epochs through a step LR scheduler. & $299\times299\times3$ & SGD LR = 0.005 momentum = 0.9 weight decay = 0.0005 & 16 & 100 epochs \\ \hline
RetinaNet \cite{retinanet} & Resnet-18-FPN used as backbone. All layers were trainable. Three output classes corresponding to normal, benign, and malignant GB. & $512\times512\times3$ & Adam LR = 0.0001 & 8 & 50 epochs \\ \hline
EfficientDet \cite{efficientdet} & EfficientNet-B4 used as backbone and BiFPN as feature network. All layers were trainable. Three output classes corresponding to normal, benign, and malignant GB. & $1024\times1024\times3$ & Adam LR = 0.001 & 2 & 50 epochs \\ \hline
MS-SoP Classifier (Ours) & 16 MS-SoP layers. All layers were trainable. Three output classes corresponding to normal, benign, and malignant GB. & $224\times224\times3$ & SGD LR = 0.005 momentum = 0.9 weight decay = 0.0005 & 16 & 100 epochs \\
\bottomrule
\end{tabular}
}
\caption{Implementation details for the different baseline networks used for classification and gallbladder localization.}
\label{tbl:configs}
\end{table*}

\section{Calculating Precision and Recall for GB Localization Networks}
\label{supp:eval_metric}
For computing precision and recall during the GB localization phase, as suggested by \cite{ribli2018detecting}, if the center of the predicted region lies within the bounding box of the ground truth region, then we consider a region prediction to be a true positive; otherwise, we consider the region prediction to be a false positive due to localization error. Further, we consider the zero/no region prediction as a false negative (all our images contain GB, and the localization network's task is to merely localize it). 

\section{GradCAM Visuals for GBCNet}
\label{supp:cam_vis}
\begin{figure*}[t]
	\centering
	\includegraphics[width=0.9\linewidth]{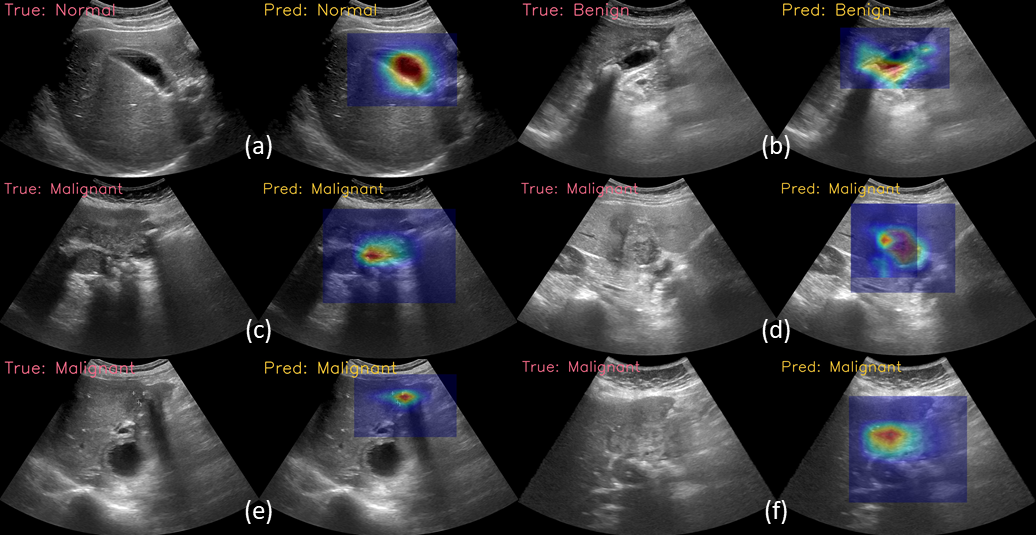}
	\caption{Sample Grad-CAM visuals of GBCNet with curriculum learning. (a) Normal, (b) Benign, and (c)--(f) Malignant samples.}
	\label{fig:supple-2}
\end{figure*}

Figure \ref{fig:supple-2} shows the sample Grad-CAM visualizations of the predictions using GBCNet (ROI+MS-SoP) with curriculum learning. 

\section{ROI Visuals}
\label{supp:roi_vis}
\begin{figure*}[t]
	\centering
	\includegraphics[width=0.9\linewidth]{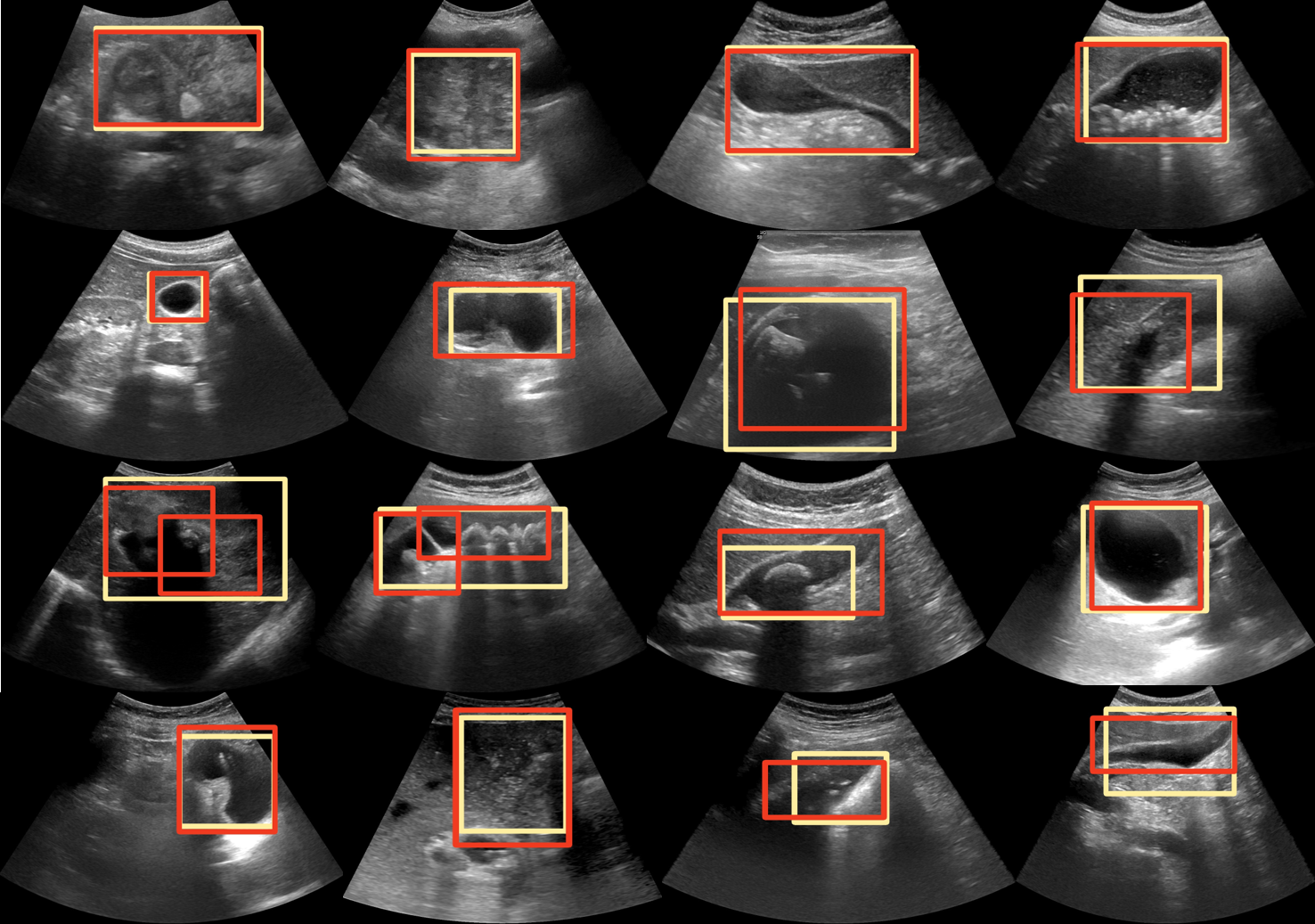}
	\caption{Sample visual results of RoI Detection models. First row - Faster-CNN, second row - YOLOv4, third row - Reppoints, and fourth row - CentripetalNet. Dark red is the ROI prediction by the model and light yellow is expert radiologists' perception of ROI.}
	\label{fig:supple-1}
\end{figure*}

In figure \ref{fig:supple-1}, we show sample predictions of the GB region localization for different models. We also show the region of interest as perceived by the expert radiologists. The localization model is fairly accurate in capturing important regions of the USG image.

\end{document}